\documentclass[letterpaper, 10 pt, conference]{ieeeconf}

\IEEEoverridecommandlockouts
\usepackage{comment}
\usepackage{cite}
\usepackage{amsmath}
\usepackage{amsfonts}
\usepackage{mathtools}
\usepackage{tabularx}
\usepackage{graphicx}
\usepackage{array}
\title{\LARGE \bf
Unsupervised Monocular Depth Learning with Integrated Intrinsics \\ and Spatio-Temporal Constraints
}

\author{Kenny Chen$^{1}$, Alexandra Pogue$^{2}$, Brett T. Lopez$^{3}$, Ali-akbar Agha-mohammadi$^{3}$, and Ankur Mehta$^{1}$%
\thanks{$^{1}$Kenny Chen and Ankur Mehta are with the Department of Electrical and Computer Engineering at the University of California Los Angeles, Los Angeles, CA 90095, USA. {\tt\small \{kennyjchen, mehtank\}@ucla.edu}}%
\thanks{$^{2}$Alexandra Pogue is with the Department of Mechanical and Aerospace Engineering at the University of California Los Angeles, Los Angeles, CA 90095, USA. {\tt\small anpogue@ucla.edu}}%
\thanks{$^{3}$Brett T. Lopez and Ali-akbar Agha-mohammadi are with the NASA Jet Propulsion Laboratory, California  Institute  of  Technology, Pasadena, CA 91109, USA. {\tt\small \{brett.t.lopez, aliagha\}@jpl.nasa.gov}}%
}

\begin{document}

\bstctlcite{IEEEexample:BSTcontrol}

\maketitle
\thispagestyle{empty}
\pagestyle{empty}


\begin{abstract}

Monocular depth inference has gained tremendous attention from researchers in recent years and remains as a promising replacement for expensive time-of-flight sensors, but issues with scale acquisition and implementation overhead still plague these systems. To this end, this work presents an unsupervised learning framework that is able to predict at-scale depth maps and egomotion, in addition to camera intrinsics, from a sequence of monocular images via a single network. Our method incorporates both spatial and temporal geometric constraints to resolve depth and pose scale factors, which are enforced within the supervisory reconstruction loss functions at training time. Only unlabeled stereo sequences are required for training the weights of our single-network architecture, which reduces overall implementation overhead as compared to previous methods. Our results demonstrate strong performance when compared to the current state-of-the-art on multiple sequences of the KITTI driving dataset and can provide faster training times with its reduced network complexity.

\end{abstract}


\section{Introduction}

Modern robotic agents take advantage of accurate, real-time range measurements to build a spatial understanding of their surrounding environments for collision avoidance, state estimation, and other navigational tasks. Such measurements are commonly retrieved via active sensors (e.g., LiDAR) which resolve distance by measuring the time-of-flight of a reflected light signal; however, these sensors are often costly \cite{royo2019overview}, difficult to calibrate and maintain \cite{katzenbeisser2003calibration, muhammad2010calibration}, and can be unwieldy for platforms with a weight budget \cite{lopez2017aggressive}. \textit{Passive} sensors, on the other hand, have seen a tremendous surge of interest in recent literature to predict scene depth from input imagery using multi-view stereo \cite{seitz2006comparison, schonberger2016pixelwise, furukawa2010towards}, structure-from-motion \cite{hartley2003multiple, newcombe2011dtam, schonberger2016structure, dellaert2000structure}, or more recently, purely monocular systems \cite{Gordon2019, Godard_digging_2019, Bian2019, Zhou2017, Yin2018, Zou2018}, due to their smaller form factor and increasing potential to rival the performance of explicit active sensors with the advent of machine learning.

In particular, monocular depth inference is attractive since RGB cameras are ubiquitous in modern times and requires the least number of sensors, but this setup suffers from a fundamental issue of scale acquisition. More specifically, in a purely monocular system, depth can only be estimated up to an ambiguous scale and requires additional geometric information to resolve the units of the depth map. Such cameras typically capture frames by projecting 3D scene information onto a 2D image plane, and abstracting higher dimensional depth information from a lower dimension is a fundamentally ill-posed problem. To resolve the scale factors of these depth maps, a variety of learning-based approaches have been proposed with differing techniques to constrain the problem geometrically \cite{Poggi2018, Godard2017, Godard_digging_2019, Garg2016, saxena2006learning, liu2015learning, saxena2008make3d, Tateno2017, Chen2019, Greene2020}. Temporal constraints, for example, are commonly employed \cite{Gordon2019, Chen2019, fu2018deep, xu2018structured} and is defined as the geometric constraint between two consecutive monocular frames, aiming to minimize the photometric consistency loss after warping one frame to the next. Spatial constraints \cite{Godard2017, Godard_digging_2019, pillai2019superdepth}, on the other hand, extract scene geometry not through a forward-backward reconstruction loss (i.e., temporally) but rather in left-right pairs of stereo images with a predefined baseline. Most works choose to design their systems around either one or the other, and while a few systems have integrated both constraints before in a multi-network framework \cite{Zhan2018, Li2018, Babu2018}, none have taken advantage of both spatial and temporal constraints in a single network to resolve these scale factors.

\begin{figure}[!t]
    \centering
    \includegraphics[width=0.85\columnwidth]{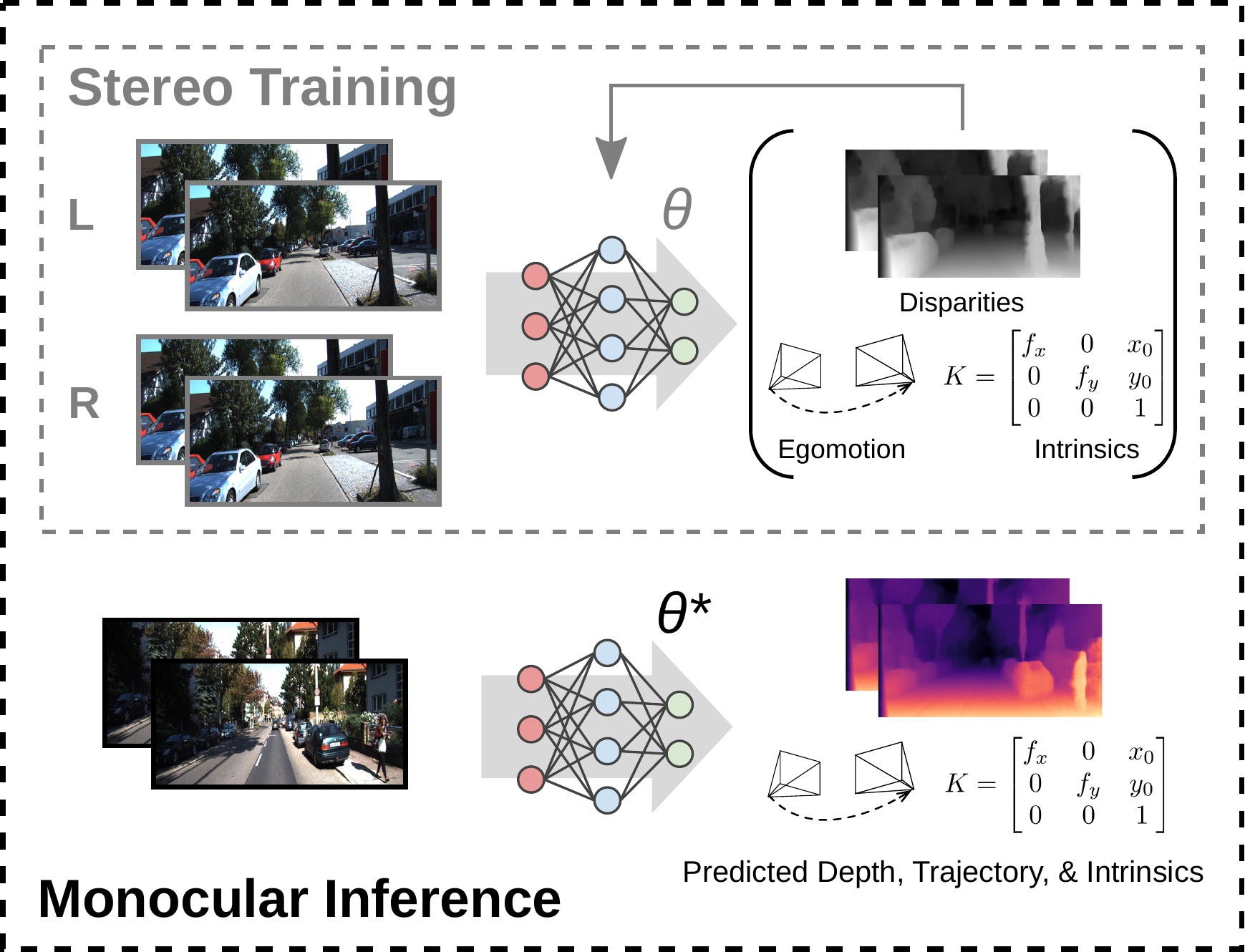}
    \vspace{-1mm}
    \caption{\textbf{System Overview.} Our system regresses depth, pose and camera intrinsics from a sequence of monocular images. During training, we use two pairs of unlabeled stereo images and consider losses in both spatial and temporal directions for our network weights. During inference, only monocular images are required as input, and our system outputs accurately scaled depth maps and egomotion in addition to the camera's intrinsics.}
    \label{fig1}
    \vspace{-5mm}
\end{figure}

\begin{figure*}[!t]
    \vspace*{3mm}
    \centering
    \includegraphics[width=0.90\textwidth]{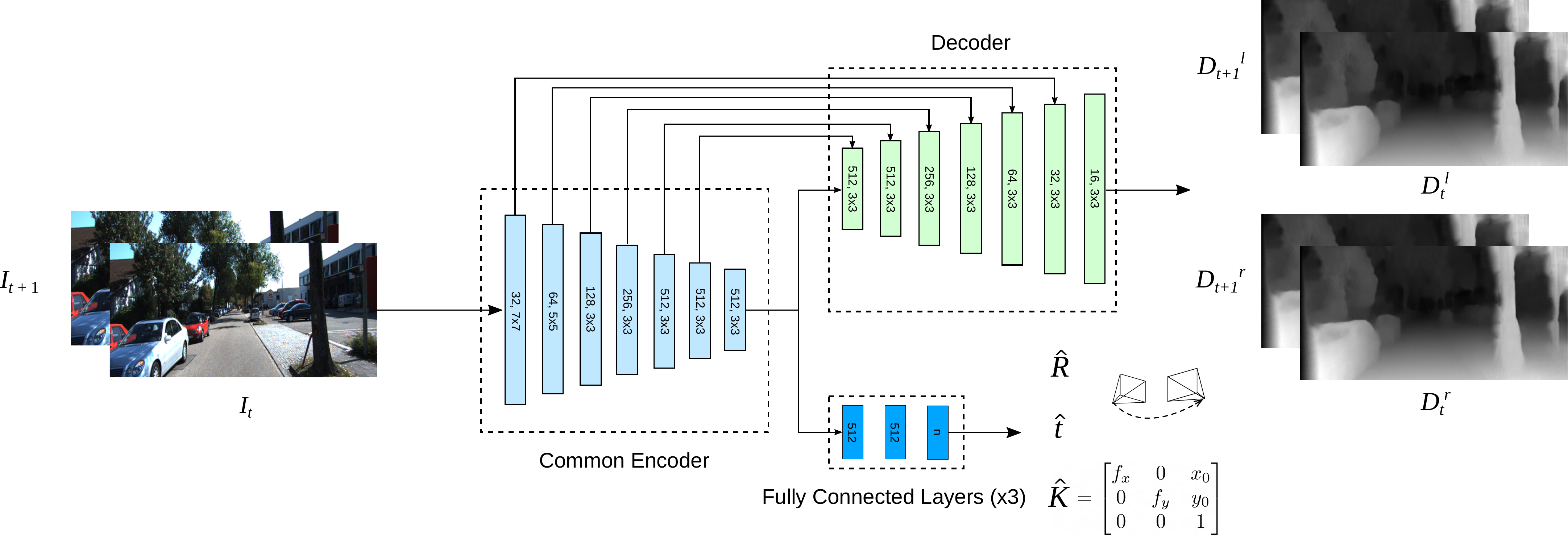}
    \vspace{-1mm}
    \caption{\textbf{Architecture Overview.} Our system uses a common convolutional-based encoder between the different outputs, which compresses the input images into a latent space representation. This representation is then sent through either a trained decoder to retrieve left-right stereo image disparities, or through different groups of fully connected layers to estimate egomotion ($n=3$) or camera intrinsics ($n=4$). In the common encoder, each block uses a series of two convolutional layers, the first with stride $2$ and the second stride $1$ (zero padding), and with input dimensions and kernel sizes as specified. The transposed convolutional blocks in the decoder are similarly structured, with pooling indices received from the corresponding encoder's feature maps.}
    \label{fig2}
    \vspace{-5mm}
\end{figure*}

To this end, we propose an unsupervised, single-network monocular depth inference approach that considers both spatial and temporal geometric constraints to resolve the scale of a predicted depth map. These ``spatio-temporal" constraints are enforced within the reconstruction loss functions of our network during training (Fig.~\ref{fig1}), which aim to minimize the photometric difference between a warped frame and the actual next frame (forward-backward) while simultaneously maximizing the disparity consistency between a pair of stereo frames (left-right). Unlike previous approaches, we consider camera intrinsics as an additional unknown parameter to be inferred and demonstrate accurate inference of both depth and camera parameters through a sequence of purely monocular frames; this is all performed in a single end-to-end network to minimize implementation overhead.

Our main contributions are as follows: (1) we propose an unsupervised, single-network architecture for monocular depth inference which takes advantage of the geometric constraints found in both spatial and temporal directions; (2) a novel loss function that integrates unknown camera intrinsics directly into the single-network training procedure; and (3) extensive performance and run-time analyses of our proposed architecture to verify our methods. These efforts were in support of NASA's Jet Propulsion Laboratory's Networked Belief-aware Perceptual Autonomy (NeBula) framework \cite{agha2021nebula} as part of Team CoSTAR in the DARPA SubT Challenge.


\subsection*{Related Work}
Depth estimation using monocular images and deep learning began with supervised methods over large datasets and  ground truth labeling \cite{Eigen_2014, eigen2015predicting, liu2015learning}. Although these methods produced accurate results, acquiring ground truth data for supervised training requires expensive 3D sensors, multiple scene views, and inertial feedback to obtain even sparse depth maps \cite{Garg2016}. Later work sought to address a lack of available high-quality labeled data by posing monocular depth estimation as a stereo image correspondence problem, where the second image in a binocular pair served as a supervisory signal \cite{Garg2016, Godard2017, Poggi2018}. This approach trained a convolutional neural network (CNN) to learn epipolar geometry constraints by generating disparity images subject to a stereo image reconstruction loss. Once trained, networks were able to infer depth using only a single monocular color image as input. While this work achieved results comparable to supervised methods in some cases, occlusion and texture-copy artifacts that arose with stereo supervision motivated learning approaches using a temporal sequence of images as an alternative \cite{Godard2017, Godard_digging_2019}. CNNs trained using monocular video regressed depth using the camera egomotion to warp a source image to its temporally adjacent target. To address the additional problem of camera pose, \cite{Zhou2017, Bian2019, Yin2018, Zou2018, Godard_digging_2019, Gordon2019, guizilini20203d, yang2018lego} trained a separate pose network. 

The learning of visual odometry (VO) and depth maps has useful application in visual simultaneous localization and mapping (SLAM). Visual SLAM leverages 3D vision to navigate an unknown area by determining camera pose relative to a constructed global map of an environment. To build and localize within a map, VO in a SLAM pipeline must solve at metric scale. Geometric approaches to monocular SLAM using first principled solutions, such as structure from motion (SfM) \cite{koenderink1991affine}, resolved scaling issues using external information \cite{clark2017vinet, Greene2020}. Building on such methods, work in data-driven monocular VO obtained scale using sources such as GPS sensor fusion \cite{pillai2017visual} or training supervision \cite{clark2017vinet, guizilini20203d}. Unsupervised approaches using a camera alone remain attractive however, due to the reduction of manual effort associated with fewer sensors. Promising research in this area combined visual constraints (e.g., monocular depth  \cite{Zhou2017, Bian2019, Yin2018, Zou2018, Godard_digging_2019, Gordon2019}, stereo depth \cite{Zhan2018, Babu2018, Li2018, Greene2020}, or optical flow \cite{Zou2018, luo2019every}) to achieve scale consistent outcomes.

Network architecture for visual odometry and dense depth map estimation separate depth and pose networks into two CNNs, one with convolutional and fully connected layers and the other an encoder-decoder structure \cite{badrinarayanan2017segnet}, respectively. In the case where only monocular images are used in training, the self-supervision inherent in estimation is less constrained, having only pose generated from temporal constraints to determine depth, and vice versa \cite{luo2019every}. The work of \cite{vijayanarasimhan2017sfmnet, Zhou2017} for example, suffered from scaling ambiguity issues\cite{Zhan2018}. Training using binocular video, on the other hand, made use of independent constraints from spatial and temporal image pairs that offered an enriched set of sampled images for network training. This ``spatio-temporal" approach allowed for the regression of depth from spatial cues generated by epipolar constraints, which were then passed to the pose network to independently estimate VO using temporal constraints\cite{Zhan2018, Babu2018, Li2018, Godard_digging_2019, luo2019every}. 

\begin{figure}[!t]
    \vspace*{3mm}
    \centering
    \includegraphics[width=0.85\columnwidth]{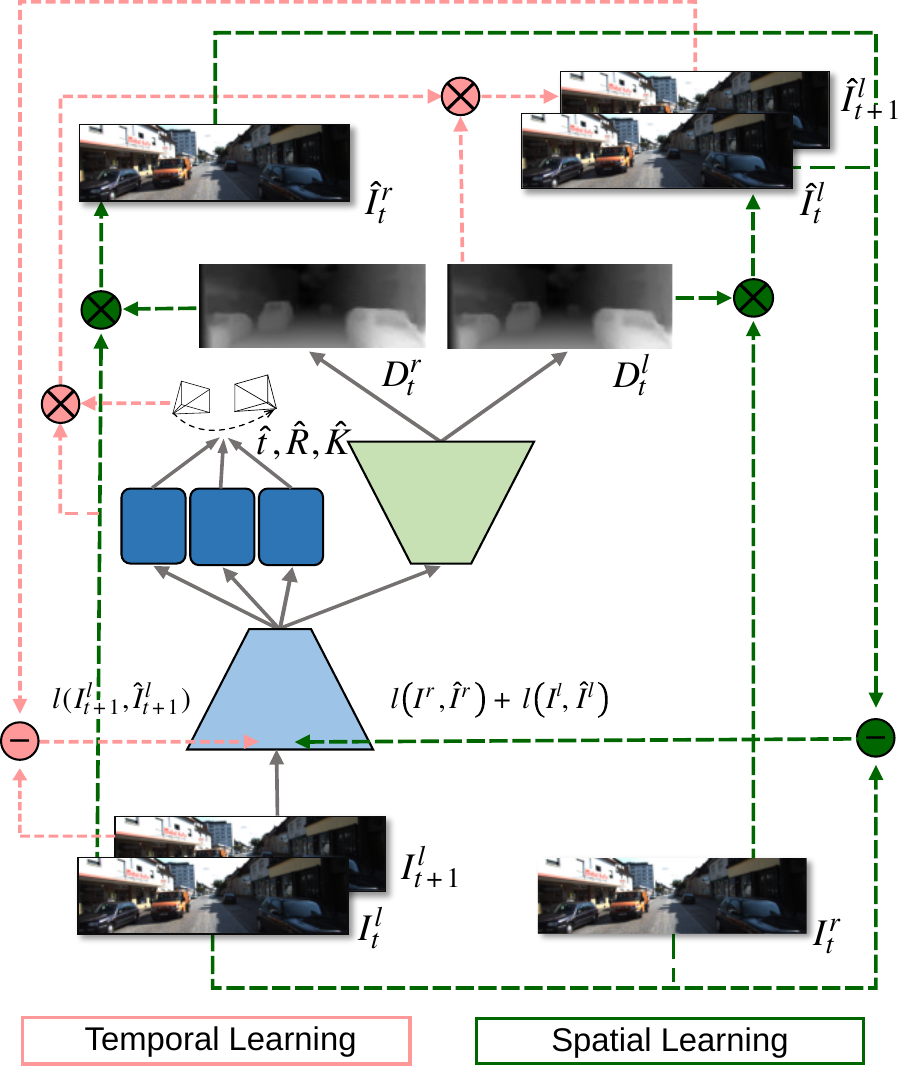}
    \caption{\textbf{Training Diagram.} Our single-network system runs a timed sequence of left images through the common encoder (light blue trapezoid) to generate outputs that are fed to the fully connected (FC) layers (blue rectangles) and the decoder (green trapezoid). Outputs from the FC layers and the decoder are the camera pose and intrinsics, and disparity maps, respectively. The disparities are used to find left-right reprojected images (green dashed lines), while the disparities, camera pose and intrinsics determine the temporal reprojections (pink dashed lines). All input and output images are framed in black for clarity.}
    \label{fig3}
    \vspace{-5mm}
\end{figure}

In this work, we propose a spatio-temporal network inspired by \cite{Zhan2018, Babu2018, Li2018} that uses an effective combination of losses to simultaneously regress depth and egomotion in a single network. Additionally, to provide freedom from manual calibration, the network is also capable of learning camera intrinsics which can be useful when a video source is unknown \cite{Chen2019, Gordon2019}. To predict depth, we use a photoconsistency loss between stereo image pairs, a left-right consistency loss between image disparity maps \cite{Godard2017}, and a disparity map smoothing function \cite{Heise_2013}. To estimate egomotion and camera intrinsics, we leverage a unique loss function that accounts for the photometric difference between temporally adjacent images. By combining these losses, we show that we can obtain scaled visual odometry information and accurate camera parameters. Furthermore, by observing the similarities between the architecture of the depth network encoder and the pose network's convolutional layers, we can effectively eliminate architecture redundancy by merging them via a common encoder (Fig.~\ref{fig2}) and can provide faster training times without significant loss in performance.


\section{Methods}
\newcommand{\hK}{\hat{K}}
\newcommand{\hR}{\hat{R}}
\newcommand{\htt}{\hat{t}}
\newcommand{\vecs}{\mathrm{v}}
\subsection{Notation}
A color image, $I$, is composed of pixels with coordinates $p_{ij} \in \mathbb{R}^2$, where $I_{ij} = I(p_{ij})$. 
In temporal training, we denote images at time $t$ as $I^t$, and images temporally adjacent as source frame $I^{t'}$. A pixel at time $t'$ is transformed to its corresponding pixel at time $t$ using homogeneous transformation matrix $T_{t'\rightarrow t} \in \mathbb{SE}(3)$ and camera intrinsics matrix $K \in \mathbb{R}^{3\times 3}$, where pixels in homogeneous coordinates, $\tilde{p} = (p,1)^T$, are denoted $p$ for simplicity.
 
Rectified stereo image pairs are given by $I^r, I^l$, where the superscripts for time have been dropped for convenience, and superscripts $l,r$ correspond to the left and right images respectively. $D^l$ represents the disparity map that warps $I^r$ to the corresponding $I^l$, and we define per pixel disparity as $d^l_{ij} = D^l(p_{ij})$. Thus $I^l_{ij} = I^r_{i+d^l, j}$, and $d^r_{i+d^l, j} = D^r(p_{i+d^l, j})$ is the disparity that does the reverse operation. Depth per pixel $z$ is then determined by the relation, $z = B f_x/ d$, where $f_x$ is the $x$-component focal length and $B$ is the horizontal baseline between stereo cameras. 
\subsection{Preliminaries}
We can obtain the projected pixel coordinates and depth map using equation, 
\begin{equation}
    z^{t} p^{t} = KR_{t'\rightarrow t}K^{-1}z^{t'}p^{t'} + Kt_{t'\rightarrow t} \,,
    \label{eq:adjacent_frame_equation}
\end{equation}
where the intrinsics matrix, $K$, is written explicitly as:  
\begin{equation}
    K = \begin{bmatrix}
    F & X_0 \\
    0 & 1
    \end{bmatrix}, \ F = diag(f_x, \ f_y), \ X_0 = [ x_0, \ y_0 ] ^T \,,
    \label{eq:intrinsic_matrix}
\end{equation}
and $R$ and $t$ are the rotation matrix and translation vector arguments of transformation matrix $T$ \cite{Gordon2019}. Note that in this work we assume no lens distortion and a zero skew coefficient in the camera, and that stereo cameras have equal intrinsic parameters. Equation  (\ref{eq:adjacent_frame_equation}) constitutes the temporal reconstruction loss at training used to determine the camera egomotion, $R$ and $t$, and the camera intrinsics $K$ in a single network. 

\subsection{Overall Optimization Objective}
\newcommand{\lnew}{l_{te}}
Our loss function is made up of a novel temporal reconstruction term and four spatial reconstruction terms \cite{Godard2017},\cite{Greene2020}. Error regression for the following losses allows the network to correctly predict a target image temporally and spatially during training in order to infer depth, pose, and camera intrinsics from a monocular image sequence at test time. The temporal reconstruction term of the loss function is implicitly defined where $\lnew (I^{l,t}, I^{l,t'}) \rightarrow \hat{I}^{l,t}$, and the spatial reconstruction terms are composed of a photoconsistency loss, $l_p$, a left-right consistency loss $l_{lr}$, and a disparity smoothness loss $l_r$,
\begin{equation}
\begin{split}
      l \left( f(I^{l}; \theta), I^{l}, I^{r} \right) &= \lambda_p \big( l_p(I^l, \hat{I}^l) + l_p(I^r, \hat{I}^r) \big) \\
                                                      &+ \lambda_{te}l_{p}(I^{l,t}, \hat{I}^{l,t} ) +  \lambda_{lr} l_{lr} \left(D^l, D^r \right) \\
                                                      &+ \lambda_r \left( l_r(D^l, I^l) + l_r(D^r, I^r) \right) \,.
\end{split}     
\label{eq:long_loss}
\end{equation}
The argument $I$ in the loss function is the original image and $\hat{I}$ is the reprojected image, and individual losses are weighted by $\lambda$ labeled with corresponding subscripts. 
\subsubsection{Spatio-Temporal Reconstruction Loss}
The \textit{photoconsistency loss} compares image appearance using the structural simularity index measure (SSIM) and an absolute error between generated and sampled images \cite{Wang_2004, Godard2017, Babu2018}:
\begin{equation}
    \begin{split}
        l_p(I, \hat{I}) = \frac{1}{N} \sum_{i,j} & \alpha \frac{1 - SSIM\left(I_{ij}, \hat{I}_{ij} \right)}{2} \\
                                                 &+ \left( 1 - \alpha \right) |I_{ij} - \hat{I}_{ij}| \,.
    \end{split}
    \label{eq:photoconsistency_loss}
\end{equation}
The loss is composed of three terms in total (two spatial losses and a temporal loss). $N$ in this equation is the number of image pixels and the weight $\alpha$ is set to $0.85$.

For reprojected images, we assume equal camera intrinsics produce right and left stereo images. The focal length $\hat{f_x}$ from instrinsics matrix $\hat{K}$ in (\ref{eq:intrinsic_matrix}) is co-predicted via the learned disparity and penalized using spatial reconstruction losses. For stereo image inputs, predicted disparity maps are used to generate the left view from a right image, and vice versa. Depth values calculated from the disparity maps are then input to the temporal reconstruction loss to generate the left target image from temporally adjacent source images, i.e. to generate the temporal image arguments for (\ref{eq:photoconsistency_loss}), we put (\ref{eq:adjacent_frame_equation}) in the form where for pixels $P = \{ p_i, i = 1 \ldots N \}$,
\begin{equation}
\begin{split}
      &\sum_{i,j} |I^{l,t}_{ij} - \hat{I}^{l,t}_{ij}| \rightarrow   \\ &\sum_{p\in P} \bigg| z^{t} p^{t} - \bigg[ \hat{K}\hat{R}_{t'\rightarrow t}\hat{K}^{-1}  \frac{b\hat{f}_x}{d^l}  p^{t'} + \hat{K}\hat{t}_{t'\rightarrow t} \bigg] \bigg| \,,
\end{split}
    \label{eq:temporal_recontruction_loss}
\end{equation}
is the absolute error between the left image and the reprojected image, and the structure similarity measure is generated by the same mappings between $I_{ij}$ and pixel $p_i$. 

We distinguish this loss function from previous works in the following ways: depth estimation in the above temporal relation is derived using spatial losses from within the same network, and the temporal loss infers both egomotion and camera intrinsics. This goes beyond work that used solely temporal constraints and a separate depth network \cite{Gordon2019}, \cite{Chen2019} or spatio-temporal work that used predetermined intrinsics and a separate depth network \cite{Godard_digging_2019}, \cite{Li2018, Babu2018, Zhan2018}, \cite{luo2019every}. 

\subsubsection{Spatial Reconstruction Loss}
The \textit{left-right disparity consistency loss} is used to obtain consistency between disparity maps \cite{Godard2017}. During training, the network predicts disparity maps $D^l$ and $D^r$ using only left image sequences as input and then penalizes the difference between the left-view disparity map and the warped right view, as well as the right-view and the warped left view,
\begin{equation}
    l_{lr}(D^l,D^r) = \frac{1}{N} \sum_{i,j} \big|d^l_{ij} - d^r_{i+d^l, j} \big| +  \big|d^r_{ij} - d^l_{i+d^r, j} \big| \,.
    \label{eq:leftright_consistency_loss}
\end{equation}
The \textit{disparity smoothness loss} penalizes depth discontinuities that occur at image gradients $\partial I$ \cite{Heise_2013}. To obtain locally smooth disparities, an exponential weighting function is used on disparity gradients $\partial d$:
\begin{equation}
    l_r(D,I) = \frac{1}{N} \sum_{i,j}\big|\partial_x d_{ij} \big| e^{-|\partial_x I_{ij}|} +
    \big | \partial_y d_{ij}| e^{-|\partial_y I_{ij}|} \,. 
    \label{eq:disparity_smoothness_loss}
\end{equation}

\subsection{Learning the Camera Intrinsics}
For predicted parameters $\hK, \ \hR, \ \htt$ in (\ref{eq:adjacent_frame_equation}), penalizing differences via training loss ensures $\hK\htt$ and $\hK^{-1}\hR\hK$ converge to the correct values. To determine parameters individually, the translational relation fails because it is under-determined since there exists incorrect values of $\hK$ and $\htt$ such that $\hK\htt = Kt$. The rotational relationship, $\hK\hR\hK^{-1} = KRK^{-1}$, however, does uniquely determine $\hK,\ \hR$ such that they are equal to $K,\ R$, and therefore provides sufficient supervisory signal to estimate these values accurately.

\begin{proof} 
From the above relation we obtain $\hR = \hK^{-1}KRK^{-1}\hK$, and we constrain $\hR$ to be $SO(3)$, i.e. $\hR^T = \hR^{-1}$ and $det(\hR) = 1$. Substituting $\hR$ into the relationship $\hR\hR^T = I$, we find that $AR = RA$ where $A = K^{-1}\hK\hK^TK^{-T}$. The value $det(\hK^{-1}KRK^{-1}\hK)$ is equal to $1$, therefore the determinant of $A$ is also equal to $1$. Moreover, the characteristic equation of $A$ shows $A$ always has an eigenvalue of $1$ \cite{Gordon2019}. Thus the eigenvalue of $A$ is equal to $1$ with an 
algebraic multiplicity of $3$, implying $A$ is the identity matrix, or the eigenvalues are unique. If we assume $A$ has $3$ distinct eigenvalues, because $A \in \mathbb{R}^{3\times3}$ and $A = A^T$, we may choose the eigenvectors of $A$ such that they are real. But because $AR = RA$, for every eigenvector, $\vecs$ of $A$, $R\vecs$ is also an eigenvector. For an eigenvalue with algebraic multiplicity $1$, the corresponding eigenspace is $dim(1)$, thus $R\vecs = \mu \vecs$ for some scalar $\mu$, implying each eigenvector of $A$ is also an eigenvector of $R$. If $R$ is $SO(3)$, however, it has complex eigenvectors in general, which contradicts this assertion. Therefore $A$ must be the identity matrix, and $\hK \hK^T = K K^T$. Referring to $K$ from (\ref{eq:intrinsic_matrix}), we observe, 
\begin{equation}
    K K^T = \begin{bmatrix} FF + X_0 X_0^T & X_0 \\ X_0^T & 1 \end{bmatrix} \,,
    \label{eq:proof_k}
\end{equation}
which implies $\hat{X}_0 = X_0$ and $\hat{F} = F$, or $\hK = K$.
\end{proof}

It is clear from above that for $R = I$, the relation $AR = RA$ holds trivially, and $\hat{K}$ cannot be uniquely determined. Thus the tolerances with which $F$ in (\ref{eq:intrinsic_matrix}) can be determined (in units of pixels) with respect to the amount of camera rotation that occurs is quantified as,
\begin{equation}
    \delta f_x < \frac{2f^2_x}{w^2r_y} \,; \quad \delta f_y < \frac{2f^2_y}{h^2r_x} \,,
\end{equation}
where $r_x$ and $r_y$ are the $x$ and $y$-axis rotation angles (in radians) between adjacent frames, and $w$ and $h$ are the width and height of the image, respectively. For a complete proof on the relation between the strength of supervision on $K$ and the closeness of $R$ to $I$, see \cite{Gordon2019}.

\subsection{Network Architecture}
Our framework is inspired by \cite{Zhan2018, Babu2018, Li2018}, but rather than requiring two separate networks for depth and pose estimation, we use a common encoder for both tasks in a novel single-network architecture (Fig.~\ref{fig3}). That is, given two temporally adjacent input images at times $t$ and $t'$, our network first convolves these inputs through a series of convolutional blocks in a common encoder, and then predicts either disparities through a decoder, or camera pose and intrinsics through fully connected layers. In the decoder network, the encoder's latent representation of the input images is re-upsampled using transposed convolutions with pooling indices from the encoder to fuse low-level features, as inspired by \cite{Godard2017, Zhan2018}. We use rectified linear units (ReLU) \cite{nair2010rectified} as activation functions in all layers of this decoder except for the prediction layer, which uses a sigmoid function instead. The decoder predicts left-to-right and right-to-left disparities $D$ at both timesteps, which are then either used to reconstruct the right stereo images for a spatially-constrained geometric loss during training via bilinear sampling, or used to construct the depth during inference. In the fully connected layers, translation $\hat{t}_{t'\rightarrow t}$, rotation $\hat{R}_{t'\rightarrow t}$, and camera intrinsics $\hat{K}$ are predicted independently in three separate and decoupled groups of fully connected layers for better performance \cite{Li2018}. These outputs are then either taken at face value during inference as the predicted egomotion and camera parameters, or used as inputs (along with the estimated depth map) to warp the current frame to the next for our temporal reconstruction loss as described previously.


\section{Results}

In this section, we evaluate our proposed framework using the KITTI driving dataset \cite{Menze2015CVPR}. Network architecture was implemented using the TensorFlow framework \cite{tensorflow2015-whitepaper} and models were trained on a single NVIDIA GeForce RTX 2070 Super GPU with 8GB of memory using a batch size of $4$. Adam optimizer \cite{kingma2014adam} was used to train the network parameters, with exponential decay rates $\beta_1 = 0.9$ and $\beta_2 = 0.99$ and learning rate $\alpha$ initially set to $0.001$ but gradually decreased throughout training. Standard data augmentation techniques were used to increase the size of the dataset during training.

We compare our method against the current state-of-the-art using conventional metrics (i.e., Abs Rel, Sq Rel, RMSE, RMSE log, and Accuracy for depth, and Absolute Trajectory Error (ATE) for egomotion) as per \cite{Zhao2020}. Table~I provides a quantitative overview of how our work fits in current literature, and Fig.~\ref{fig4} accompanies this table with a visual, qualitative comparison. Tables~II~and~III evaluate odometry and camera intrinsics estimation performance, and Fig.~\ref{fig5} shows a comparison of trajectory on the KITTI Odometry dataset. We additionally evaluate our framework's run-time performance to show the benefits of our reduced network size and overhead while maintaining comparable performance with current methods which can be seen in Fig.~\ref{fig6}.

\subsection{Performance of Depth Estimation}

To evaluate the performance of our network's depth inference, we use a standard Eigen split \cite{Eigen_2014} on the KITTI dataset \cite{Menze2015CVPR} as per convention and compare against several state-of-the-art methods with a depth cap of $80m$. We train, validate, and test our network using these splits and compare our depth estimation accuracy against multiple other works across several metrics, as shown in Table~I. Ground truth data for the testing set is calculated via projecting the Velodyne LiDAR data onto the image plane.

\begin{table*}[!t]
    \setlength{\tabcolsep}{12 pt}
    \renewcommand{\arraystretch}{1.2}
    \vspace{5mm}
    
    TABLE I: Comparison of monocular depth estimation with state-of-the-art approaches. Cropped regions from \cite{Godard2017} were used for performance evaluation all methods. In the column labeled ``Type", ``D" indicates supervised training with ground truth, ``T" indicates temporal training only, ``S" indicates spatial training only, and ``ST" indicates a spatio-temporal training approach. Column ``A.C." denotes whether additional components (such as pre-/post-processing methods) were used in addition to neural networks, and column ``Comb." denotes whether pose and depth networks were combined. Column ``Int." denotes whether that method can simultaneously regress camera intrinsics \{Yes (Y), No (N)\} (a dash ``-" indicates no applicability). We evaluate using the Eigen split \cite{Eigen_2014} on the KITTI dataset \cite{Menze2015CVPR} and cap depth to $80m$ as per standard practice \cite{Godard2017}. Results from other methods were taken from their corresponding papers. For error metrics, lower is better; for accuracy, higher is better.
    
    \begin{center}
    \begin{tabular}{|p{18mm}>{\centering\arraybackslash}m{1mm}>{\centering\arraybackslash}m{2mm}>{\centering\arraybackslash}m{2mm}>{\centering\arraybackslash}m{2mm}||>{\centering\arraybackslash}m{4mm}>{\centering\arraybackslash}m{4mm}>{\centering\arraybackslash}m{4mm}>{\centering\arraybackslash}m{8mm}|>{\centering\arraybackslash}m{10mm}>{\centering\arraybackslash}m{10mm}>{\centering\arraybackslash}m{10mm}|}
    \hline
    \multicolumn{5}{|c||}{} & \multicolumn{4}{c|}{Error Metrics} & \multicolumn{3}{c|}{Accuracy Metrics} \\
    Method & Type & A.C. & Comb. & Int. & Abs~Rel & Sq~Rel & RMSE & \mbox{$\textrm{RMSE}_{log}$} & \mbox{$\delta<1.25$} & \mbox{$\delta<1.25^2$} & \mbox{$\delta<1.25^3$} \\
    \hline
    Train~Set~Mean                                      & D & - & - & - & 0.361 & 4.826 & 8.102 & 0.377 & 0.638 & 0.804 & 0.894 \\
    Zou~\textit{et~al.}~\cite{Zou2018}                  & T & \textbf{Y} & N & N & 0.150 & 1.124 & 5.507 & 0.223 & 0.806 & 0.933 & 0.973\\
    Yin~\textit{et~al.}~\cite{Yin2018}                  & T & \textbf{Y} & N & N & 0.149 & 1.060 & 5.567 & 0.226 & 0.796 & 0.935 & 0.975 \\
    Chen~\textit{et~al.}~\cite{Chen2019}                & T & N & N & \textbf{Y} & 0.135 & 1.070 & 5.230 & 0.210 & 0.841 & 0.948 & 0.980 \\
    Gordon~\textit{et~al.}~\cite{Gordon2019}            & T & \textbf{Y} & N & \textbf{Y} & 0.128 & 0.959 & 5.230 & - & - & - & - \\
    Guizilini~\textit{et~al.}~\cite{guizilini20203d}    & T & \textbf{Y} & N & N & \textbf{0.111} & \textbf{0.785} & \textbf{4.601} & \textbf{0.189} & \textbf{0.878} & \textbf{0.960} & \textbf{0.982} \\
    Garg~\textit{et~al.}~\cite{Garg2016}                & S & N & - & N & 0.177 & 1.169 & 5.285 & 0.282 & 0.727 & 0.896 & 0.958 \\
    Godard~\textit{et~al.}~\cite{Godard2017}            & S & N & - & N & 0.148 & 1.344 & 5.927 & 0.247 & 0.803 & 0.922 & 0.964 \\
    Poggi~\textit{et~al.}~\cite{Poggi2018}              & S & N & - & N & 0.163 & 1.399 & 6.253 & 0.262 & 0.759 & 0.911 & 0.961 \\
    Pillai~\textit{et~al.}~\cite{pillai2019superdepth}  & S & N & N & N & 0.116 & 0.935 & 5.158 & 0.210 & 0.842 & 0.945 & 0.977 \\
    Luo~\textit{et~al.}~\cite{luo2019every}             & ST & \textbf{Y} & N & N & 0.127 & 0.936 & 5.008 & 0.209 & 0.841 & 0.946 & 0.979 \\
    Godard~\textit{et~al.}~\cite{Godard_digging_2019}   & ST & \textbf{Y} & N & N & 0.127 & 1.031 & 5.266 & 0.221 & 0.836 & 0.943 & 0.974 \\
    Li~\textit{et~al.}~\cite{Li2018}                    & ST & N & N & N & 0.183 & 1.730 & 6.570 & 0.268 & - & - & - \\
    Babu~\textit{et~al.}~\cite{Babu2018}                & ST & N & N & N & 0.139 & 1.174 & 5.590 & 0.239 & 0.812 & 0.930 & 0.968 \\
    Zhan~\textit{et~al.}~\cite{Zhan2018}                & ST & N & N & N & 0.144 & 1.391 & 5.869 & 0.241 & 0.803 & 0.928 & 0.969 \\
    \textbf{Ours}                                       & ST & N & \textbf{Y} & \textbf{Y} & 0.141 & 1.227 & 5.629 & 0.239 & 0.809 & 0.927 & 0.962 \\
    \hline
    \end{tabular}
    \end{center}
    \label{table:results_with_error_metric}
    \vspace{-1mm}
\end{table*}

\begin{table*}[!t]
    \setlength{\tabcolsep}{12 pt}
    \renewcommand{\arraystretch}{1.2}
    
    TABLE II: Comparison of our system's odometry estimation against various other state-of-the-art methods \cite{Babu2018, Zhou2017, geiger2011stereoscan} using absolute trajectory error for translation ($t_{ate}$) and rotational ($r_{ate}$) movement. Comparison was done on four sequences of the KITTI dataset.
    
    \begin{center}
    \begin{tabular}{|c||cc|cc|cc|cc|cc|cc|}
    \hline
    Seq. & \multicolumn{2}{c|}{Ours} & \multicolumn{2}{c|}{UnDEMoN \cite{Babu2018}}  & \multicolumn{2}{c|}{SfMLearner \cite{Zhou2017}} & \multicolumn{2}{c|}{VISO-M \cite{geiger2011stereoscan}}\\
     & $t_{ate}$ & $r_{ate}$ & $t_{ate}$ & $r_{ate}$ & $t_{ate}$ & $r_{ate}$ & $t_{ate}$ & $r_{ate}$ \\
    \hline
    00 & 0.0712 & 0.0014 & \textbf{0.0644} & 0.0013 & 0.7366 & 0.0040 & 0.1747 & \textbf{0.0009} \\
    04 & \textbf{0.0962} & 0.0016 & 0.0974 & \textbf{0.0008} & 1.5521 & 0.0027 & 0.2184 & 0.0009 \\
    05 & \textbf{0.0689} & \textbf{0.0009} & 0.0696 & \textbf{0.0009} & 0.7260 & 0.0036 & 0.3787 & 0.0013 \\
    07 & 0.0753 & 0.0013 & \textbf{0.0742} & \textbf{0.0011} & 0.5255 & 0.0036 & 0.4803 & 0.0018 \\
    \hline
    
    \end{tabular}
    \end{center}
    \label{table:egomotion_comparison}
    \vspace{-1mm}
\end{table*}

\begin{figure*}[!t]
    \centering
    \includegraphics[width=0.85\textwidth]{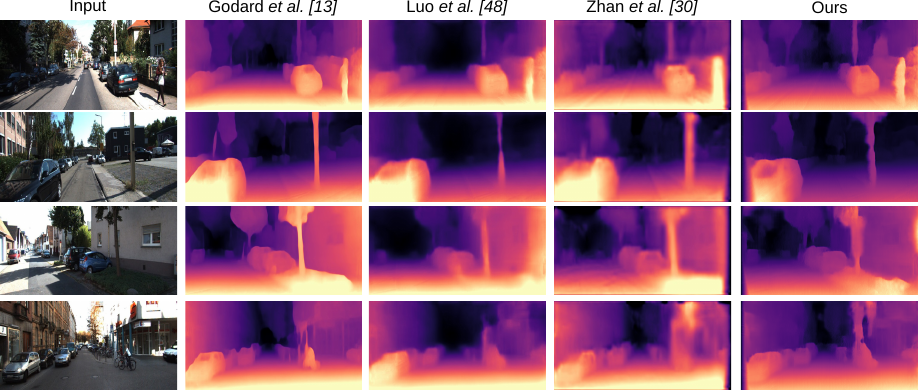}
    \caption{\textbf{Qualitative Comparison of Depths.} Visual comparison of regressed depth maps between our method and various state-of-the-art methods (\cite{Godard_digging_2019, luo2019every, Zhan2018}) on four images from the KITTI Eigen split (images of other methods were retrieved from \cite{Godard_digging_2019}). Even with a reduced size and complexity, our network can accurately regress depth maps given a single monocular image.}
    \label{fig4}
\end{figure*}

Several important observations can be extracted from Table~I. First, and most notably, our method is the only one which provides a combined architecture for both pose and depth regression (as indicated in the ``Comb." column). Whereas all previous methods split egomotion and depth estimation tasks into separate pipelines, we reduce network redundancy (and hence, number of parameters needed for training) by sharing the input latent representation for both tasks via a common encoder. A second observation is that only two other works before ours are also capable of estimating camera intrinsics during the inference phase (\cite{Chen2019, Gordon2019}, as indicated in the ``Int." column). From these observations, our proposed method, to the best of our knowledge, is the first to demonstrate a simultaneous regression of pose, depth, and camera parameters in a reduced single-network design that uses both spatial and temporal geometric constraints. 

However, this reduction in network complexity may come with trade-offs. First, through a quantitative lens, Table~I shows that our method is not the lowest in depth estimation error or highest in accuracy. With a reduced network complexity, a possible explanation lies in our encoder's shared network parameters that must balance both depth and pose/intrinsics pathways. However, even then, our error and accuracy is still strongly comparable to many state-of-the-art methods, many of which have additional components to compensate for occlusions, motion, etc. Compared to the current best with the lowest error and highest accuracy \cite{guizilini20203d}, we are on average ${\sim}75.9\%$ as error-free and ${\sim}95.6\%$ as accurate. This can also be seen qualitatively in Fig.~\ref{fig4}. Our method lacks slightly in sharpness compared to \cite{Godard_digging_2019} but can provide finer edges than \cite{luo2019every} and \cite{Zhan2018}. Depending on one's setup, the benefits of faster training through a more compact network could outweigh such trade-offs.

\subsection{Learned Camera Intrinsics}

To evaluate our system's ability to recover camera intrinsics (i.e., $f_x, f_y$, $x_0, y_0$) through the supervisory signal provided by the rotational component of (\ref{eq:adjacent_frame_equation}), we follow a similar procedure as \cite{Gordon2019} and trained separate models on several different video sequences until convergence of these parameters for multiple independent results. During training, parameters were randomly initialized to begin with and empirically had no convergence issues throughout our experiments. We used ten video sequences of the ``2011\_09\_28" subdataset chosen to have the same ground truth calibration done that day, and Table III shows the resulting mean and standard deviation of those ten tests. All experiments were done on the left stereo color camera (``image\_02") of the vehicle setup. For all four variables, we observe that, on average across all ten tests, our method can learn the parameters accurately and within a reasonable bound.

\begin{table}[!t]
    \renewcommand{\arraystretch}{1}
    \vspace{5mm}
    TABLE III: Regressed camera intrinsics during training as compared to the ground truth. Note that ground truth values have been adjusted to match the scaling and cropping done for training. All values are in units of pixels.
    
    \begin{center}
    \begin{tabular}{lcc}
    Camera Parameter & Learned & Ground Truth \\
    \hline
    Horizontal Focal Length ($f_x$) & 298.4 $\pm$ 2.3 & 295.8 \\
    Vertical Focal Length ($f_y$) & 483.1 $\pm$ 3.6 & 489.2 \\
    Horizontal Principal Point ($x_0$) & 254.8 $\pm$ 2.4 & 252.7 \\
    Vertical Principal Point ($y_0$) & 127.8 $\pm$ 1.7 & 124.9
    \end{tabular}
    \end{center}
    \label{table:intrinsics}
    \vspace{-8mm}
\end{table}

\subsection{Egomotion}

\begin{figure}[!t]
    \centering
    \vspace{4mm}
    \includegraphics[width=0.60\columnwidth]{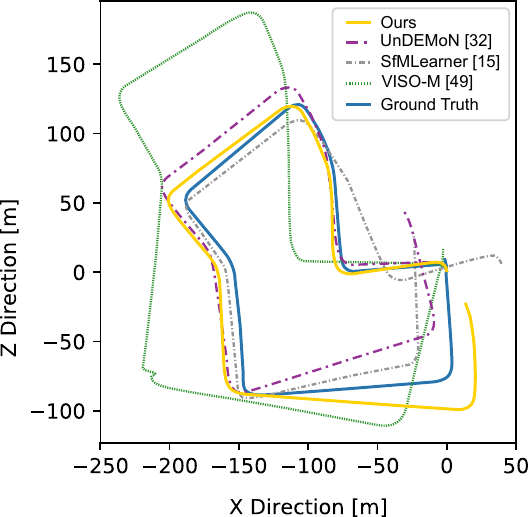}
    \vspace{-2mm}
    \caption{\textbf{Trajectory Comparison.} Visual comparison of estimated egomotion between our method and several others (\cite{Babu2018, Zhou2017, geiger2011stereoscan}) on Sequence $07$ of the KITTI Odometry dataset. Corresponding $t_{ate}$ and $r_{ate}$ metrics can be found in Table~II.}
    \label{fig5}
    \vspace{-1mm}
\end{figure}

We carried out our pose estimation performance evaluation using four sequences from the KITTI Odometry dataset \cite{geiger2012we} and compared against several state-of-the-art methods, including UnDEMoN \cite{Babu2018}, SfMLearner \cite{Zhou2017}, and VISO-M \cite{geiger2011stereoscan}. For a quantitative comparison, we adopt the absolute trajectory root-mean-square error (ATE) for both translational ($t_{ate}$) and rotational ($r_{ate}$) components per standard practice \cite{sturm2012benchmark}. We note that we used the same model that was trained for depth estimation to output our egomotion estimation, and that these four test sequences were not part of our training set. Sequence $07$ is shown in Fig.~\ref{fig5}.

From Table II, we observe that for both translational and rotational errors in all four sequences, our method outperformed SfMLearner \cite{Zhou2017} and VISO-M \cite{geiger2011stereoscan} and is comparable with UnDEMoN's \cite{Babu2018} performance. In contrast to these methods, our system co-predicts egomotion and camera intrinsics (alongside disparity) in a single network such that the loss functions for these free parameters are tied together. This may explain the slight loss in accuracy, especially when compared to \cite{Babu2018}, but the upside is that our method is a reduction in computational complexity as there are fewer weights in our architecture to optimize over.

\subsection{Run-Time Evaluation}

Our single-network architecture design via a common encoder decreases overall network complexity (and therewith the number of network parameters) which can decrease the necessary time to optimize weights. Previously, when using a 7-layer CNN instead of our common encoder for pose and intrinsics regression, network size was ${\sim}30$ million in trainable parameters; however, after replacing those layers with a common encoder, the number of trainable parameters reduced to ${\sim}27.6$ million. This is around an $8\%$ savings.

\begin{figure}[!t]
    \centering
    \includegraphics[width=0.60\columnwidth]{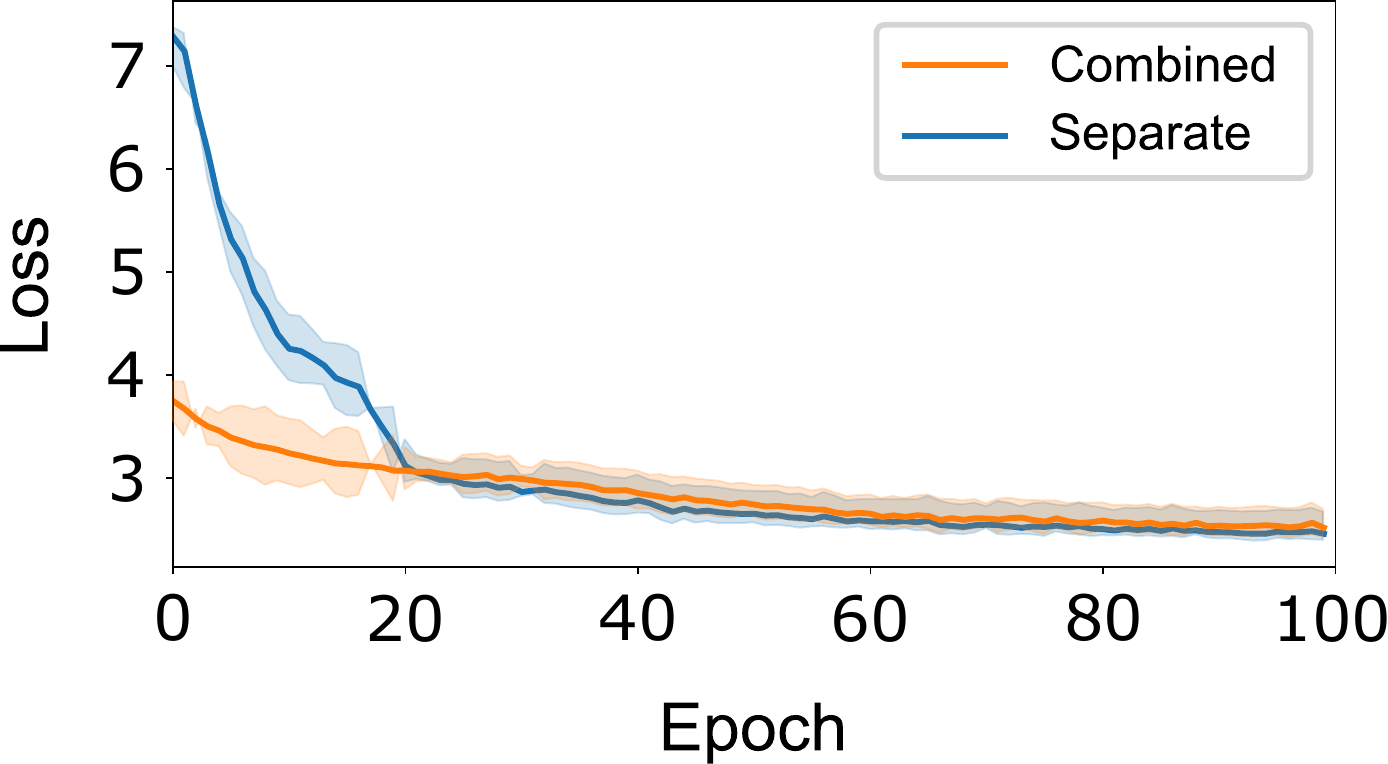}
    \vspace{-2mm}
    \caption{\textbf{Loss Comparison.} Average training loss of the first $100$ epochs for both architecture variants. The solid lines represent the mean across ten different runs, and the shaded areas represent one standard deviation (${\sim}70\%$ confidence). The ``Separate" architecture used a 7-layer CNN for pose and intrinsics, while ``Combined" used a common encoder.}
    \label{fig6}
    \vspace{-6mm}
\end{figure}

Fig.~\ref{fig6} shows the effects of this decrease through a run-time comparison. In this figure, the mean (solid) and standard deviation (shaded) loss for each epoch is calculated across ten independent runs for each model. We observed a smaller initial average loss in the combined network, where in this case initialization of weights from the 7-layer CNN that may contribute to a higher loss before being optimal is not necessary. Over time though, we observed that these losses cross paths (roughly at $20$ epochs), which is likely caused by the additional $2.4$ million parameters for its function approximation. Thus, for those looking to maximize accuracy, a separated network may be better; however, for others who need reasonable results very quickly, a combined network can provide that in just a few epochs.


\section{Discussion}

In this work we have presented an unsupervised, single-network monocular depth inference approach for joint prediction of environmental depth, egomotion, and camera intrinsics. Through training our neural network to learn spatial and temporal constraints between stereo and temporally-adjacent pairs, we are able to resolve solutions at metric scale using only monocular video at test time. We distinguish our work from other monocular inference approaches by creating a single, fully differentiable architecture for depth prediction and visual odometry. To further reduce human effort and manual intervention, we also take advantage of intrinsics observability in the system by learning the camera parameters embedded within the temporal reconstruction loss. We verify the success of our system using the KITTI dataset, where our results show we are able to achieve performance comparable to the state-of-the-art in monocular vision while solving for intrinsics and decreasing overhead and overall training complexity.

In future work we plan to quantify network robustness to initialization error during the training of camera parameters. We are also interested in comparing depth and odometry results between predetermined and learned intrinsics, to analyze their effects on prediction outcomes. To improve performance, we expect that the expansion of training to other datasets will provide a more diverse collection of scenes for evaluation, and learned intrinsics will allow for pooling of datasets as another avenue for training. Additionally addressing occlusion and moving objects will ensure added support for higher complexity scenes captured within these datasets.

\bibliographystyle{IEEEtran}
\bibliography{IEEEbib}

\end{document}